\documentclass[letterpaper, 10 pt, conference]{ieeeconf}  

\IEEEoverridecommandlockouts                              

\usepackage{amsmath,amsfonts}
\usepackage{algorithmic}
\usepackage{array}
\usepackage[caption=false,font=normalsize,labelfont=sf,textfont=sf]{subfig}
\usepackage{textcomp}
\usepackage{stfloats}
\usepackage{url}
\usepackage{verbatim}
\usepackage{graphicx}
\usepackage{booktabs}
\usepackage{threeparttable}
\usepackage[linesnumbered, ruled, vlined, boxed, commentsnumbered]{algorithm2e}
\usepackage{cite}
\usepackage{hyperref} 
\hypersetup{
    colorlinks=true,
    linkcolor=black,
    citecolor=black, 
    filecolor=magenta,      
    urlcolor=blue,
    }

\usepackage{amsmath,amssymb,amsfonts}
\usepackage{textcomp}
\usepackage{adjustbox}
\usepackage[table]{xcolor}

\begin{document}
\setlength{\headsep}{5pt}
\title{\LARGE \bf BEV-LIO(LC): BEV Image Assisted LiDAR-Inertial Odometry with Loop Closure
\vspace{-3mm}
}
\author{
Haoxin Cai$^1$, Shenghai Yuan$^2$, Xinyi Li$^1$, Junfeng Guo$^1$, and Jianqi Liu$^{*1}$
\vspace{-4.5mm}
\thanks{$^1$Authors with the School of Computer Science and Technology,
Guangdong University of Technology, Guangzhou 510006, China. \texttt{\small hxcai@mail2.gdut.edu.cn, liujianqi@ieee.org} ($^*$Corresponding author: Jianqi Liu).
}
\thanks{$^2$Shenghai Yuan is with the School of Electrical and Electronic Engineering, Nanyang Technological University 63979, Singapore. \texttt{\small shyuan@ntu.edu.sg}.
}
\thanks{This work was supported in part by the National Science Foundation of China under Grant 62172111, National Joint Fund Key Project (NSFC - Guangdong Joint Fund) under
Grant U21A20478.
\vspace{-0.5mm}}%
}





\maketitle
\begin{abstract}
This work introduces BEV-LIO(LC), a novel LiDAR-Inertial Odometry (LIO) framework that combines Bird's Eye View (BEV) image representations of LiDAR data with geometry-based point cloud registration and incorporates loop closure (LC) through BEV image features. By normalizing point density, we project LiDAR point clouds into BEV images, thereby enabling efficient feature extraction and matching. A lightweight convolutional neural network (CNN) based feature extractor is employed to extract distinctive local and global descriptors from the BEV images. Local descriptors are used to match BEV images with FAST keypoints for reprojection error construction, while global descriptors facilitate loop closure detection. Reprojection error minimization is then integrated with point-to-plane registration within an iterated Extended Kalman Filter (iEKF). In the back-end, global descriptors are used to create a KD-tree-indexed keyframe database for accurate loop closure detection. When a loop closure is detected, Random Sample Consensus (RANSAC) computes a coarse transform from BEV image matching, which serves as the initial estimate for Iterative Closest Point (ICP). The refined transform is subsequently incorporated into a factor graph along with odometry factors, improving the global consistency of localization. Extensive experiments conducted in various scenarios with different LiDAR types demonstrate that BEV-LIO(LC) outperforms state-of-the-art methods, achieving competitive localization accuracy. Our code and video can be found at \url{https://github.com/HxCa1/BEV-LIO-LC}.
\end{abstract}


\vspace{1.5mm}
\section{Introduction}
Recent advances in LIO have greatly enhanced the efficiency and accuracy of simultaneous localization and mapping (SLAM). Methods like FAST-LIO2 \cite{FAST-LIO2} have demonstrated exceptional performance, making the fusion of LiDAR and inertial sensors a popular choice for odometry.
\par However, the sparsity of point clouds in LiDAR SLAM systems presents challenges. Unlike image data with a structured pixel grid, the irregular and sparse distribution of point clouds in 3D space complicates the extraction of stable keypoints and unique features, potentially resulting in reduced localization precision. While approaches like \cite{LIVO, LIC-Fusion2.0} fuse visual information with LiDAR data to address these challenges, they require additional sensors, precise extrinsic calibration, and time synchronization, yet struggle in low-light scenarios. Some works project point clouds into range images to improve the performance, such as MD-SLAM \cite{MD-SLAM} which relies on multi-cue image pyramids generated from range images, encoding surface normals and intensity information for pose-graph optimization. While range image-based methods are robust to view rotations due to the equivalence between point cloud rotation and horizontal image shifting, they suffer from scale distortions caused by spherical projection, limiting the accuracy of localization.

\par BEV representations, as demonstrated in several works \cite{BEVPlace, BEVPlace++, BVMatch}, offer a promising alternative by projecting LiDAR point clouds into structured 2D images. This approach has gained popularity in loop closure detection \cite{ScanContext, LiDARB}, as BEV preserves scale consistency and spatial relationships, enabling robust feature extraction via CNNs. Unlike spherical range images, BEV avoids scale distortion and generalizes across different LiDAR types, making it particularly suitable for global localization and loop closure detection. However, existing BEV-based methods primarily focus on global tasks, such as place recognition and loop detection, while neglecting tight integration with real-time odometry frameworks. This limits their ability to fully exploit BEV's potential for LiDAR odometry and its extension to robust loop closure detection in SLAM.
\begin{figure}[!t]
\centering
\includegraphics[width=\linewidth]{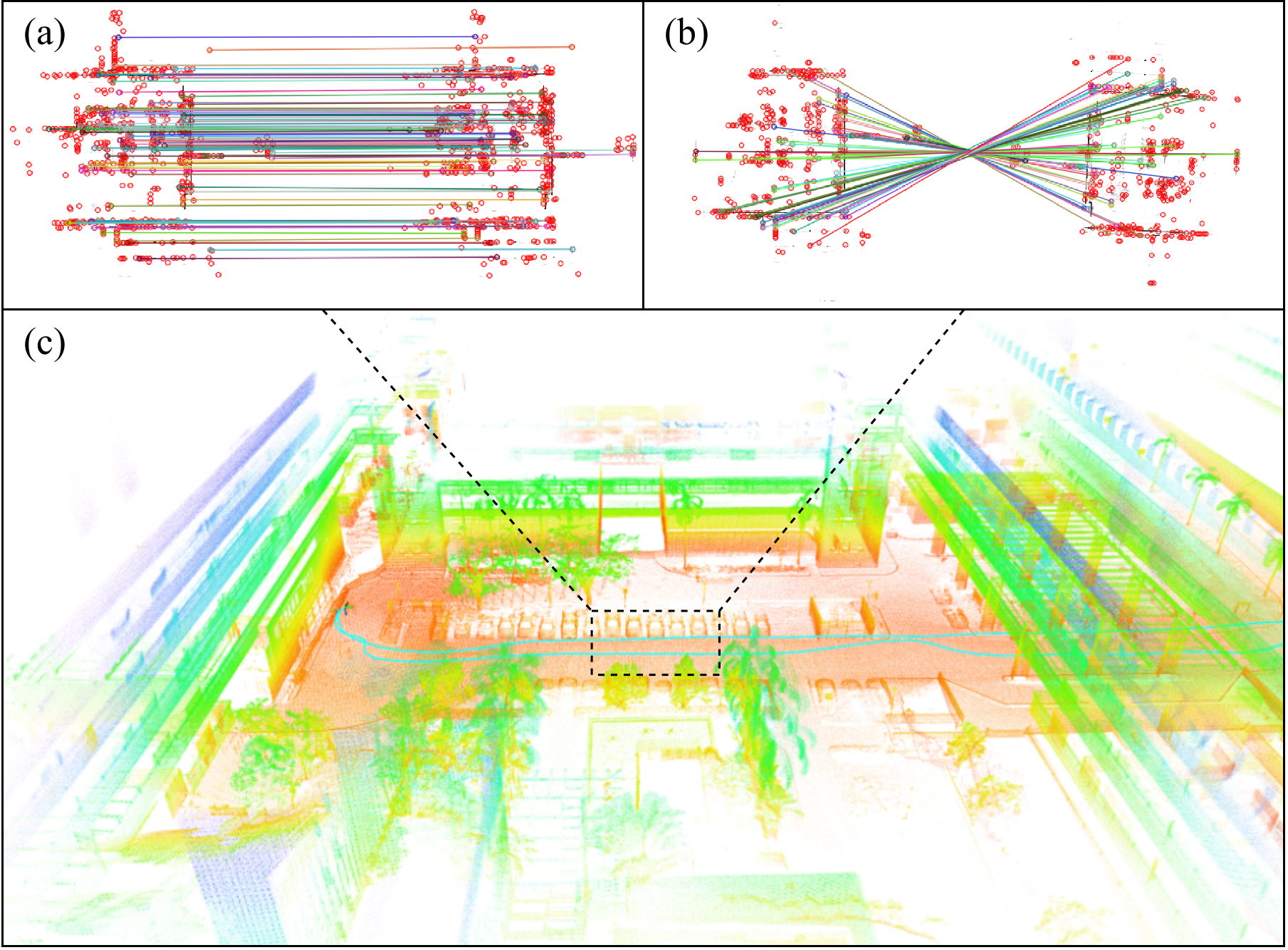}
\caption{BEV-LIO(LC) utilizes BEV image features to perform frame-to-frame matching, constructing reprojection residuals coupled with geometric residuals, thereby enhancing the accuracy of loop closure implementation. The upper left image (a) shows BEV feature frame-to-frame matching (the matching lines are downsampled by 90\% for better visualization and clarity), the upper right image (b) shows loop image pair matching, and (c) presents the map constructed using our method.}
\label{introduction}
\vspace{-7.5mm}
\end{figure}

\par Therefore, we present BEV-LIO(LC), a LIO framework that utilizes BEV images to tightly couple geometric registration with reprojection error minimization and robust loop closure detection (Fig. \ref{introduction}). The key contributions of our work are as follows:
\begin{itemize}
    \item We integrate BEV feature reprojection errors with geometric registration via analytic Jacobian derivation, establishing a tightly-coupled iEKF framework that improves front-end odometry accuracy.
    \item We propose a loop closure algorithm that utilizes a KD-tree-indexed keyframe database with global descriptors for efficient candidate retrieval. Detected candidates first undergo RANSAC-based coarse alignment using BEV image matching, then ICP refinement between geometric measurements. The optimized transformation is integrated into a factor graph with odometry factors to improve the global consistency of localization.
    \item Extensive experimental results demonstrate that BEV-LIO(LC) outperforms state-of-the-art methods in various environments with different LiDAR types.

\end{itemize}


\vspace{-2mm}
\section{Related Works}
\vspace{-0.5mm}
\subsection{LiDAR (Inertial) SLAM}
\vspace{-1mm}

LiDAR-based SLAM has long been significantly influenced by LOAM \cite{LOAM}, which splits the SLAM problem into two parallel tasks: odometry and mapping. By extracting edge and planar features from LiDAR scans and optimizing feature correspondences over long-term optimization, LOAM achieves high accuracy with low computational cost in structured environments. This approach has inspired subsequent works such as LeGO-LOAM \cite{LeGO-LOAM} and LIO-SAM \cite{LIO-SAM}. LeGO-LOAM introduced lightweight ground-optimized segmentation and loop closure mechanisms. LIO-SAM innovated through Inertial Measurement Unit (IMU) tight-coupling in factor graphs \cite{FactorGraph} with the Bayes tree \cite{ISAM2}, enabling motion-aware submap matching and drift-aware loop closure integration. However, traditional methods like \cite{LOAM} and its variants struggle in featureless environments or with LiDARs that have a small field of view. To address these challenges, FAST-LIO \cite{FAST-LIO} introduces an iEKF update mechanism, enabling real-time alignment of each scan with an incrementally built map. In its later iteration, FAST-LIO2, based on the ikd-Tree, replaces feature matching with point-to-plane ICP on raw points, offering improved robustness and state-of-the-art performance across various environments. DLIO \cite{DLIO} further utilizes analytical equations for fast and parallelizable motion correction. By directly registering scans to the map and employing a nonlinear geometric observer, DLIO improves both accuracy and computational efficiency. LTA-OM \cite{LTAOM} integrates FAST-LIO2 for front-end odometry and STD \cite{STD} for loop closure, further incorporating loop optimization. Its multisession operation enables dynamic map updates and robust localization, yielding performance gains over state-of-the-art SLAM systems. iG-LIO \cite{ig-lio} introduces a tightly-coupled LIO framework using GICP, a voxel-based surface covariance estimator that replaces kd-tree-based method \cite{GICP}, \cite{VGICP} to accelerate processing in dense scans, and an incremental voxel map, enhancing efficiency while maintaining state-of-the-art accuracy. Recently, correspondence-based methods employ geometric primitives and robust estimation to reduce problem size and prune correspondences \cite{TEASER, Quatro}, yet point feature degradation and rigid assumptions yield unreliable matches and unresolved non‑planar ambiguities \cite{Segregator, Outram}. Although these methods achieve competitive performance in efficiency and accuracy, they exhibit inherent limitations in diverse environments due to their over-reliance on geometric features.

\vspace{-1mm}

\subsection{Other Information Assisted LiDAR (Inertial) SLAM}
\vspace{-1mm}
In addition to LiDAR SLAM, several methods have been proposed to enhance robustness by incorporating additional information. I-LOAM \cite{I-LOAM} and Intensity-SLAM \cite{I-SLAM} integrate intensity as a similarity metric, incorporating it into a weighted ICP approach. Similarly, approaches like \cite{I-LOAM,I-SLAM} use high-intensity points as an additional feature class for more accurate registration. \cite{MD-SLAM} focuses on optimizing photometric errors using intensity, range, and normal images, but doesn't incorporate IMU data or perform motion undistortion. RI-LIO \cite{RI-LIO} integrates reflectivity images within a tightly-coupled LIO framework by leveraging photometric error minimization into the iEKF of \cite{FAST-LIO}, aiming to efficiently reduce the drift in geometric-only methods. COIN-LIO \cite{COIN-LIO} improves LIO by integrating LiDAR intensity with geometric registration, utilizing a novel image processing pipeline and feature selection strategy for enhanced robustness in geometrically degenerate environments, such as tunnels. By selecting complementary patches and continuously assessing feature quality, it performs well in these scenarios. However, its focus on geometrically degenerate environments (e.g., long corridors) limits its adaptability, resulting in compromised performance in more diverse environments. Critically, these methods either rely on dense-channel imaging LiDARs for reliable operation or suffer from scale distortions caused by spherical projection, which limits their localization accuracy and restricts their applicability to a broader range of LiDAR configurations, particularly for sparse-channel LiDAR systems. Therefore, we need another better source of information to assist LiDAR SLAM. 

\vspace{-1.5mm}
\subsection{Related BEV Approaches}
\vspace{-1mm}
Recent advances in LiDAR localization, place recognition and loop closure have explored BEV representations to improve accuracy.
MapClosures \cite{MapClosures} proposes a loop closure detection method for SLAM utilizing BEV density images with ORB features derived from local maps, enabling effective place recognition and detection of map-level closures. 
\cite{BVMatch} pioneers BEV-based place recognition by projecting 3D LiDAR scans to BEV images, generating rotation-invariant maximum index maps using Log-Gabor filters, and employing the novel Bird’s-Eye View Feature Transform (BVFT) for robust feature extraction and pose estimation. 
BEVPlace \cite{BEVPlace} and its extension \cite{BEVPlace++} further advance this concept using lightweight CNNs with rotation-equivariant modules and NetVLAD \cite{NetVLAD} global descriptors, achieving state-of-the-art performance in subtasks of global localization including place recognition and loop closure detection.
\par In contrast to direct point cloud matching or spherical projection based methods, BEV representations inherently avoid scale distortions by projecting 3D points into a unified 2D plane, thereby enabling stable improvements in SLAM performance across different LiDAR configurations as demonstrated in our experiments. Thereby, we propose BEV-LIO(LC), a novel LIO framework that leverages BEV features to tightly integrate geometric registration, reprojection error minimization, and loop closure.
\vspace{-0.5mm}

\begin{figure*}[!t]
   \vspace{-0.5mm}
\centering
    \vspace{-1.5mm}
\includegraphics[width=\linewidth]{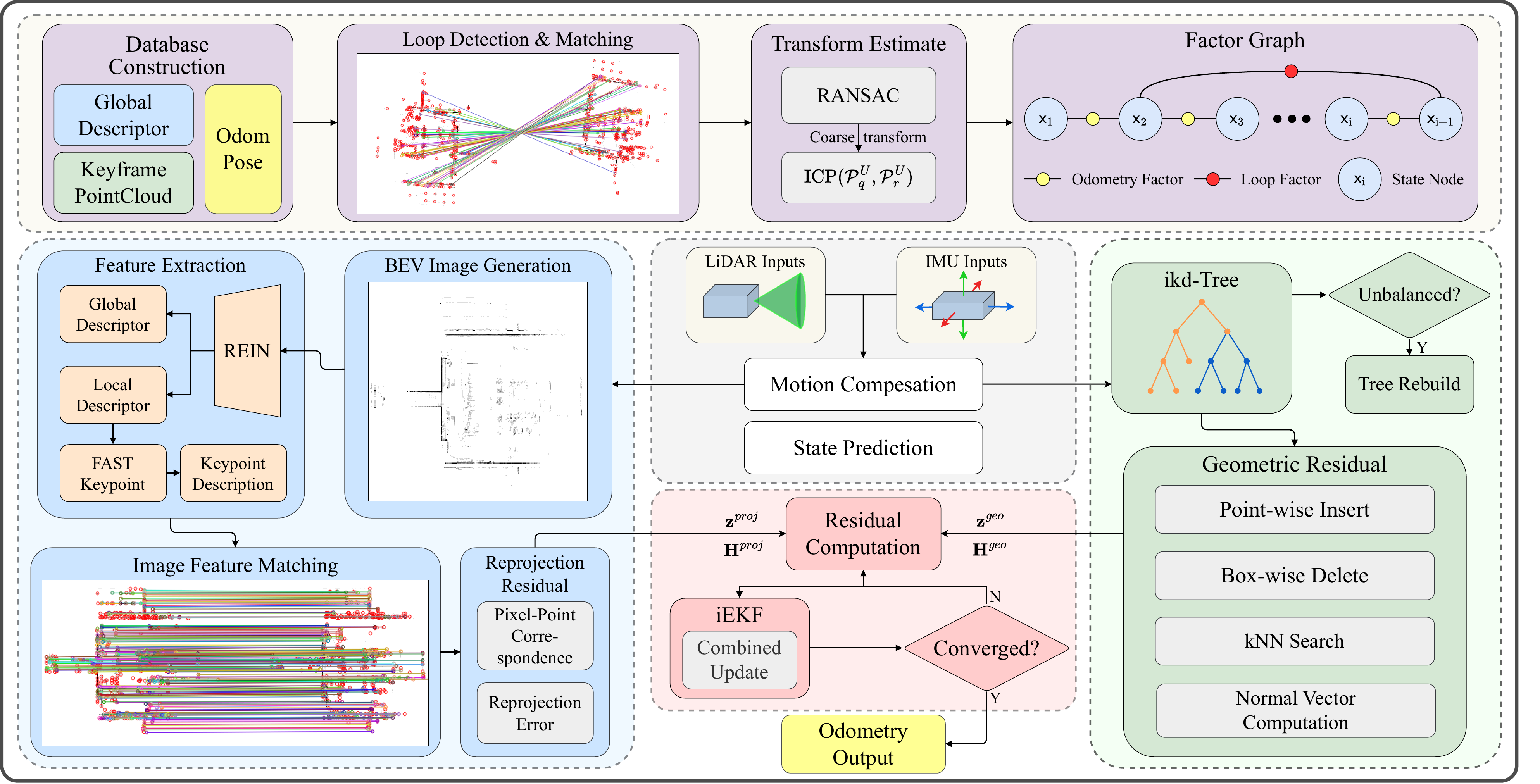}
\caption{
System overview of BEV-LIO and BEV-LIO-LC. The system first preprocesses LiDAR scans with IMU motion compensation before constructing geometric and reprojection residuals. Geometric residuals (green, right bottom) are computed via ikd-Tree with kNN search and normal vector computation. Reprojection residuals (blue, left bottom) are derived from BEV image feature matching and constructed by pixel-point correspondence. Both residuals are fused within the iEKF for state estimation (red). To mitigate drift, BEV-LIO-LC incorporates a loop closure module (purple), which detects and verifies loops via global descriptor and BEV image matching and refines constraints in a factor graph.}
\label{system overview}
    \vspace{-5.5mm}
\end{figure*}

    \vspace{-0.5mm}

\section{Method}
    \vspace{-0.5mm}

BEV-LIO(LC) builds upon the tightly-coupled iEKF framework originally proposed in FAST-LIO2 for point-to-plane registration. While COIN-LIO extends FAST-LIO2 with photometric error minimization, BEV-LIO introduces a novel approach by incorporating reprojection error minimization by BEV images matching. Since FAST-LIO2 has been widely studied, we focus on the components of BEV-LIO and BEV-LIO-LC. As shown in Fig. \ref{system overview}, our method begins with motion prediction and point cloud undistortion to correct LiDAR measurements. The undistorted point cloud is then projected into a BEV image by normalizing point density. We adopt the Rotation Equivariant and Invariant Network (REIN) \cite{BEVPlace++} to extract distinct local and global descriptors from the BEV image. Local descriptors with keypoints are matched to construct the reprojection error, which measures the alignment between the current frame and the keyframe. Global descriptors are used for loop closure detection by querying a keyframe database constructed in real-time. Finally, the reprojection residual is integrated into the iEKF framework to refine the pose estimation.
\vspace{-1mm}

\subsection{Symbol Definition}
We denote the world frame as $(\cdot)_W$, the LiDAR frame as $(\cdot)_L$ and the IMU frame as $(\cdot)_I$. Transformation from LiDAR frame to IMU frame is represented as $\mathbf{T}_{IL} = (\mathbf{R}_{IL},\mathbf{p}_{IL}) \in SE(3)$. 
The robot state $\mathbf{x} = [\mathbf{R}_{WI}, _W\mathbf{p}_{WI},_W\mathbf{v}_I, \mathbf{b}_a,\mathbf{b}_g, \mathbf{g}]$
at the $i$-th LiDAR frame is defined as $\mathbf{x}_i$, where $\mathbf{R} \in SO(3)$ denotes the orientation, $\mathbf{p} \in \mathbb{R}^3$ represents the position, $\mathbf{v} \in \mathbb{R}^3$ is the linear velocity, $\mathbf{b}_g \in \mathbb{R}^3$ and $\mathbf{b}_a \in \mathbb{R}^3$ indicate the gyroscope and accelerometer biases, respectively, and $\mathbf{g} \in \mathbb{R}^3$ denotes the gravity vector.
\par Each LiDAR scan from a full revolution can be represented as a point set $\mathcal{P}=\{P_{i}\}^{N_{p}}_{i=1} $, where $P_i=[x_i,y_i,z_i]$ denotes the $i$-th LiDAR point, and $N_p$ is the total number of points.
\vspace{-1mm}
\subsection{IMU Prediction and point cloud undistortion}

We adopt the Kalman Filter prediction step according to FAST-LIO2 [1] by propagating the state using IMU measurement integration from $t_{i-1}$ to $t_i$. Similarly, we calculate the ego-motion compensated, undistorted points at the latest timestamp $t_i$ as: 
$P_i= 
\mathbf{T}_{L_i I_i} 
\mathbf{T}_{I_i I_{i-1}} 
\mathbf{T}_{I_{i-1} L_{i-1}} P_{i-1}$. Then the undistorted point set can be described as $\mathcal{P}_{i}^U$.
\vspace{-1mm}

\subsection{BEV Image Projection Model}\label{projection}
According to existing 3-DoF localization works \cite{ScanContext, MapClosures,RIL}, we assume that the ground vehicle moves on a rough plane within a local area. Following the assumption, we project the LiDAR scans orthogonally to BEV image and focus on optimizing pose estimation in 3-DoF, including (x, y, yaw).
\par We use the normalized point density to construct BEV images. For a scan $\mathcal{P}$, we first downsample the point cloud using a voxel grid filter with a leaf size of $g$ meters. Then we discretize the ground plane into a 2D Cartesian grid $N_i(u,v)\in\mathbb{N}_0^{v_i\times u_i}$ with a resolution of $\mu$ meters. Considering a $[2 y_{max} , 2 x_{max}]$ rectangle window centered at the coordinate origin, each point $P_i = [x_i,y_i,z_i]$ projected in the image can be described as:
\vspace{-0.5mm}
\begin{equation}
\Pi(P_i) = 
\left[
\begin{array}{c}
u_i \\
v_i
\end{array}
\right]
= 
\left[
\begin{array}{c}
\displaystyle \lfloor\frac{y_{\text{max}} - y_i} {\mu} \rfloor \\
\displaystyle \lfloor\frac{x_{\text{max}} - x_i}{\mu} \rfloor
\end{array}
\right]
\end{equation}
\vspace{-0.5mm}
\par Each cell in the grid $N_i(u_i,v_i)$ stores the point count per cell after discretization and $(u_i,v_i)$ indicates the distribution of $\mathcal{P}$ in  $\mathbb{R}^{2}$. The BEV intensity $\mathbf{I}(u,v)$ is defined as:
\vspace{-1mm}
\begin{equation}
    \mathbf{I}(u,v)=\frac{min(N_i,N_m)}{N_m}
\end{equation}
where $N_m$ denotes the normalization factor that is set as the max value of the point cloud density.

\par Our approach utilizes point densities to construct the BEV image, diverging from traditional BEV projection methods \cite{ScanContext, ScanContext++, SPI, LiDARB} that rely on storing the maximum height of the point cloud to build an elevation map. While elevation maps are effective in preserving local 2D geometry and reducing computational complexity, they are inherently sensitive to sensor pose changes, as the recorded maximum height varies with the scanner's distance to objects. In contrast, our method captures 2.5D structural information of the environment by focusing on point densities, which exhibit lower sensitivity to viewpoint variations. By disregarding point distribution along the z-axis, our BEV image retains the rigid structures on the x-y plane, providing a more robust representation of the egocentric environment.


\subsection{Feature Extraction \& Matching}\label{Feature}
\begin{figure}[!tb]
\centering
\includegraphics[width=\linewidth]{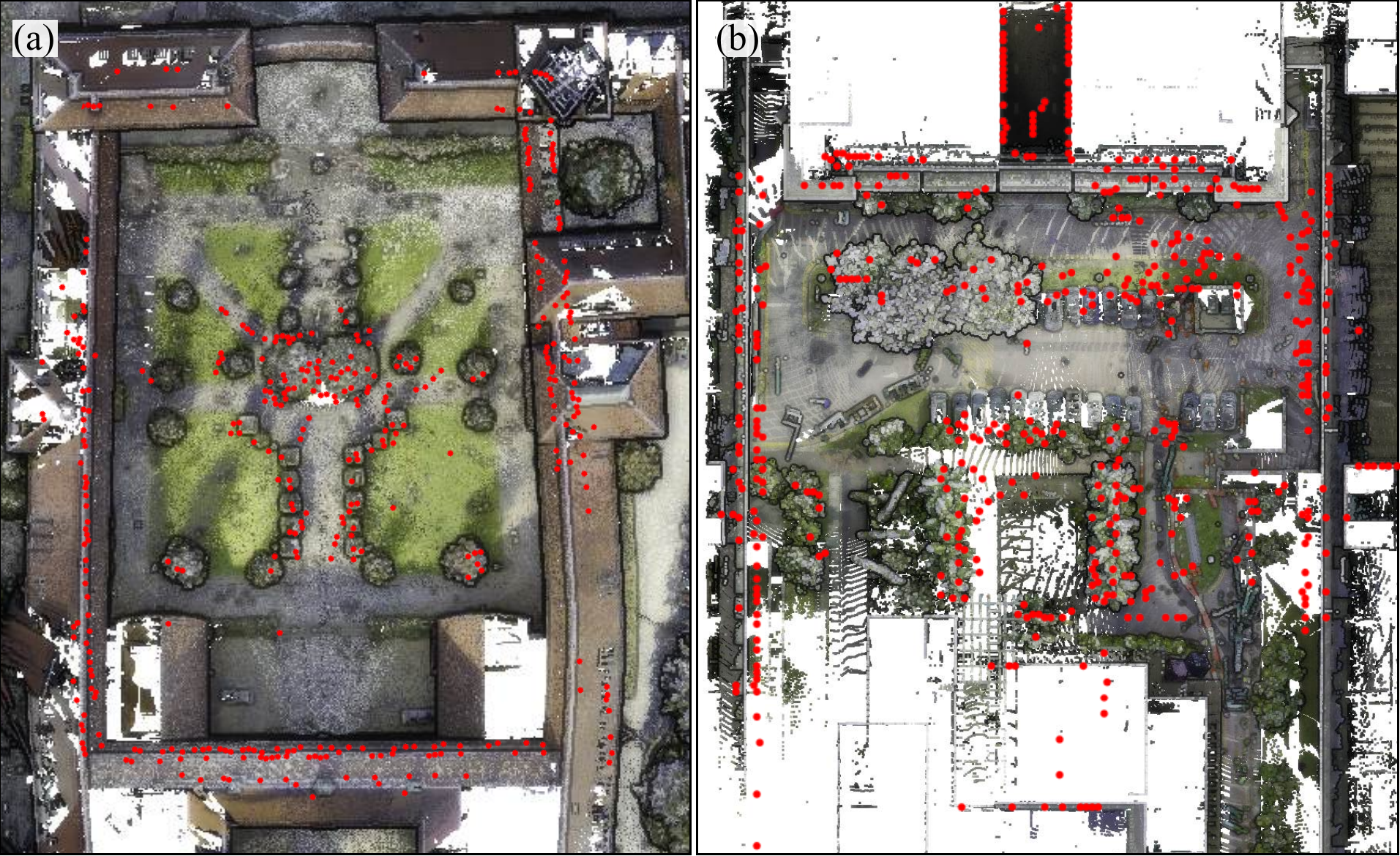}
\caption{Examples of BEV image features. They highlight BEV image features from one scan of mobile low-cost LiDAR (red) distributed across the Leica prior map of KTH (a) and NTU (b) campus, both conforming to geometrically consistent structures.}
\label{BEVImage}
\vspace{-6.5mm}
\end{figure}
 \par \textbf{1) Feature Extraction.} For feature extraction, we adopt the REIN \cite{BEVPlace++} to generate rotation-equivariant local descriptors for front-end odometry and viewpoint-invariant global descriptors for back-end loop closure detection. Given a BEV image $\mathbf{I} \in \mathbb{R}^{H \times W}$, the rotation equivariant module outputs a feature map $\mathbf{F} \in \mathbb{R}^{H' \times W' \times C}$ with dimension parameters $(H', W', C)$ {denoting} height, width, and feature channels respectively. The local descriptors are mainly used in frame-to-frame matching, enabling computation of a reprojection residual. To facilitate loop closure detection, these local descriptors are aggregated by NetVLAD in the REIN to generate a global descriptor $\mathbf{V} \in \mathbb{R}^{K \times C}$ where $K$ is the number of clusters in NetVLAD. The loop closure will be introduced in Section \ref{Loop Closure}.
 \par \textbf{2) BEV Image Feature Matching.} We start by extracting FAST keypoints \cite{FAST} from the BEV images for fast and accurate detection. As demonstrated in Fig. \ref{BEVImage} and \ref{BEVLoopImage}, these keypoints exhibit strong repeatability, as they often correspond to vertical structures in the environment (e.g., facades, poles, signposts). Each keypoint is then assigned a local descriptor interpolated from the REIN. Using these local descriptors and the keypoints, we perform feature matching between pairs of BEV images as Fig. \ref{introduction} (a) does and optimize the feature matching with RANSAC to construct a reprojection error model.

\vspace{-1.5mm}

\subsection{Reprojection Residual \& Kalman Update}\label{Reprojection Residual}
We minimize frame-to-frame reprojection errors using the matched features. The error is computed by projecting the matched features in current frame into the last frame. Knowing that feature points $p_i$ and $p_{i-1}$ in the BEV images correspond to the projection of the same spatial point $P$, we calculate the reprojection error as the difference between the projected position of the 3D point and its observed location. The error can be described as:
\vspace{-0.5mm}
\begin{equation}
\label{eq:project}
\mathbf{z}^{proj} = 
\left[
\begin{array}{c}
u_i \\
v_i
\end{array}
\right] - \mathbf{T}
\left[
\begin{array}{c}
u_{i-1} \\
v_{i-1}
\end{array}
\right]
\end{equation}
where $(u_i,v_i)$ and $(u_{i-1},v_{i-1})$ are calculated by feature points $p_i$ and $p_{i-1}$.

\par Following the description in eq. (\ref{eq:project}), we then calculate the change in the projected image coordinates due to a perturbation of the point in a given direction using eq. (\ref{eq:H}):
\vspace{-0.5mm}
\begin{equation}
\label{project_H} 
\mathbf{d}_{p_i} = 
\frac{\partial{\Pi(P_i})}
        {\partial{P_i}}
\end{equation}
\vspace{-1mm}
The resulting Jacobian $H^{proj}$ is computed as follows:
\vspace{-0.5mm}
\begin{equation}
\label{eq:H}
    \mathbf{H}^{proj}_i = \eta  \cdot
    \frac{\partial{\Pi(P_i)}}{\partial{P_i}} \cdot
    \frac{\partial{P_i}}{\partial{\tilde{\mathbf{x}}}}
\end{equation}

\vspace{-3mm}

\begin{equation}
\begin{aligned}
    \frac{\partial P_i}{\partial \tilde{\mathbf{x}}} 
    &= (\mathbf{R}_{L_i L_{i-1}} \mathbf{R}_{LI}) \\
    &\quad  \begin{bmatrix} 
        \left[ -\mathbf{R}_{IW} ({}^W P_i - {}^W \mathbf{p}_{WI}) \right]_\times & 
        -\mathbf{R}_{IW} & 
        \mathbf{0}
    \end{bmatrix}
\end{aligned}
\end{equation}
where $\eta$ is a factor that scales the projection transformation to maintain consistency across different directions. 
\par Similar to \cite{COIN-LIO}, we then combine the point-to-plane terms ($^{geo}$) with the reprojection terms ($^{reproj}$) into a unified residual vector ($\mathbf{z}$) and Jacobian matrix ($\mathbf{H}$). And the parameter $\alpha$ balances the discrepancy in error magnitudes associated with the geometric and reprojection residuals:
\vspace{-1mm}
\begin{align*}
    &\mathbf{H} = \left[\mathbf{H}^{geoT}_1, \cdots, \lambda \cdot \mathbf{H}^{projT}_1, \cdots, \lambda \cdot \mathbf{H}^{projT}_n\right]^T \\
    \mathbf{z}^{\kappa}_k &= 
    \left[
    \mathbf{z}^{geo}_1, \cdots, \mathbf{z}^{geo}_m, \lambda \cdot \mathbf{z}^{proj}_1, \cdots, \lambda \cdot \mathbf{z}^{proj}_n
    \right]^T 
    \mathbf{R} = \operatorname{diag}[\alpha]    
\end{align*}
The formulas provided in \cite{FAST-LIO2} is used to update the state:

\vspace{-1mm}
\begin{equation}
        \mathbf{K} = 
        (\mathbf{H}^T\mathbf{R}^{-1}\mathbf{H}
        +\mathbf{P}^{-1})^{-1}
        \mathbf{H}^T\mathbf{R}^{-1}
\end{equation}
\vspace{-3mm}
\begin{equation}
        \widehat{\mathrm{x}}^{\kappa+1}_k
        =\widehat{\mathrm{x}}^{\kappa}_k \boxplus (-\mathbf{Kz^{\kappa}_k}-
        (\mathbf{I}-\mathbf{KH})
        (J^{\kappa})^{-1} (\widehat{\mathrm{x}}^{\kappa}_k \boxminus \widehat{\mathrm{x}}_k)
        )
\end{equation}

\subsection{Loop Closure}\label{Loop Closure}
Our back-end adopts a factor graph framework FAST-LIO-SAM\cite{FAST-LIO-SAM}, integrating loop closure constraints through the following pipeline:

\par \textbf{1) Keyframe Database Construction.} The database incrementally stores keyframe tuples containing:
\begin{equation}
\mathcal{D}=\{ (\mathcal{P}^U_i, \mathbf{T}^G_i, \mathbf{V}_i) \}_{i=1}^n
\end{equation}
where $\mathcal{P}^U_i$ is the undistorted point cloud, $\mathbf{T}^G_i$ the global pose, and $\mathbf{V}_i$ the NetVLAD global descriptor extracted from $\mathcal{P}_{i}^U$ of keyframe $i$.


\begin{figure}[!tb]
\centering
\includegraphics[width=\linewidth]{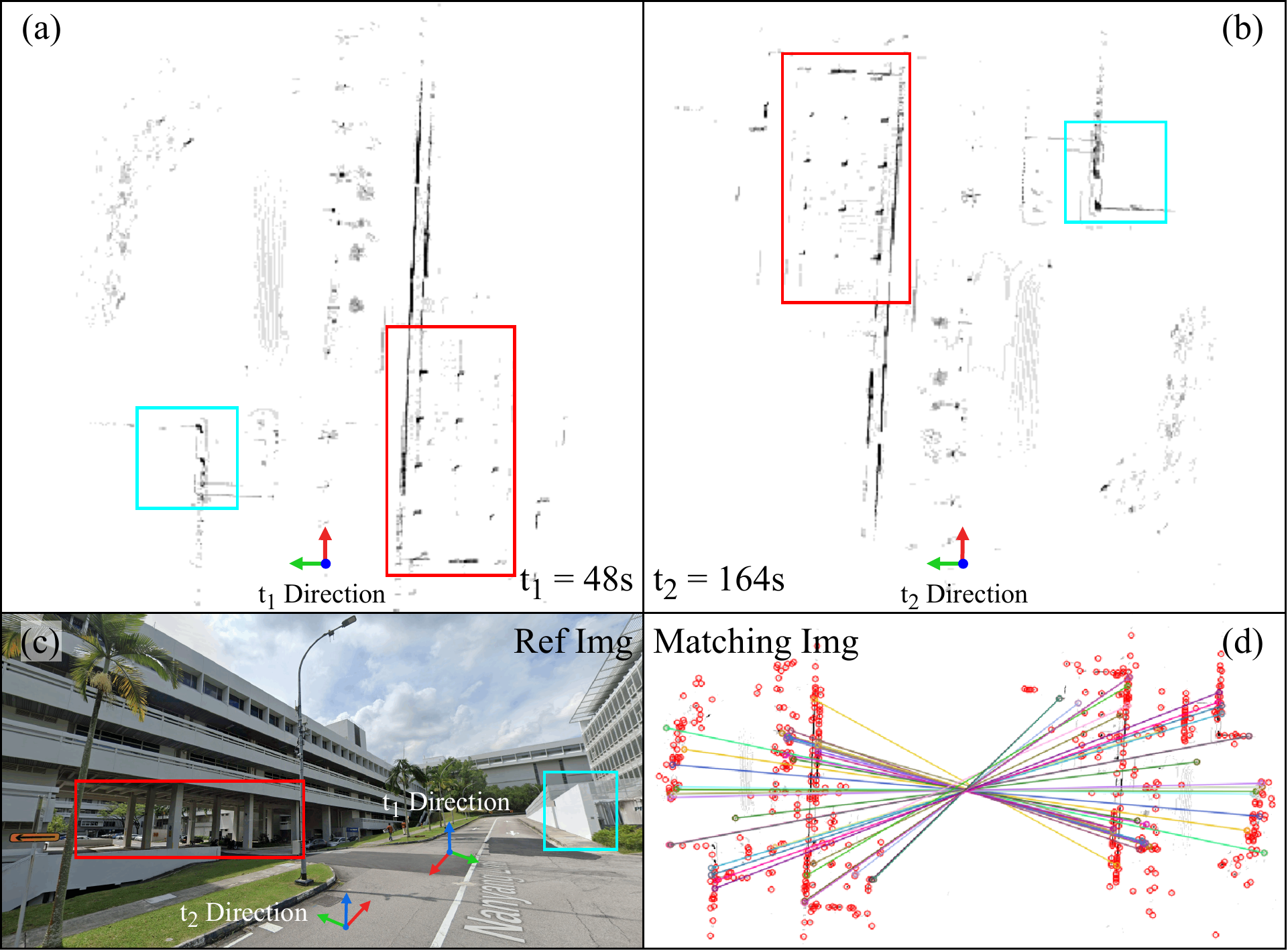}
\caption{An example of loop closure: (a) BEV image, (b) Current BEV image, (c) Reference image of NTU, (d) Feature matching result.}
\label{BEVLoopImage}
\vspace{-6.5mm}
\end{figure}

\par \textbf{2) Loop Detection.} The global descriptors $\mathbf{V}$ in the database $\mathcal{D}$ are indexed by a KD-tree for efficient retrieval of candidate keyframes during loop detection. For a query descriptor $\mathbf{V}^Q$, we employ the L2 distance metric to search for the $k$-nearest global descriptors $\mathbf{V}^{R}$ and their corresponding indices in $\mathcal{D}$:  
\vspace{-1mm}
\begin{equation}
\mathbf{dist} = \arg\min_{j=1,2,...,n} \|\mathbf{V}^{Q} - \mathbf{V}_j\|_2
\end{equation}  
\vspace{-5mm}

The $k$ descriptors with the smallest distances are added to the candidate set $\mathbf{V}^{R}$. If the Euclidean distance between the pose $\mathbf{T}^G_j$ associated with $\mathbf{V}^{R}_j$ and the query pose $\mathbf{T}^{G}_q$ is less than a predefined threshold $\tau$, a loop closure is detected. The index of the current keyframe is denoted as $Q$ or $q$, and the index of the reference loop keyframe is denoted as $R$ or $r$.

\par \textbf{3) Coarse Alignment by BEV Image.} As discussed in Section \ref{projection} and \ref{Feature}, we generate BEV images from undistorted point clouds stored in $\mathcal{D}$ and extract FAST keypoints with local descriptors. As Fig. shown in \ref{BEVLoopImage}, feature matching is then performed between reference BEV image (a) and current BEV image (b) to establish correspondences, which are used to estimate a coarse relative transformation between the BEV images pair by RANSAC. 
\par Since the BEV images are orthogonally projected from point clouds, the transformation between BEV images is rigid. Therefore, the transformation $\mathbf{T}_{rq}$ between the corresponding point clouds can be recovered by the transformation of BEV image pairs. A rotation matrix $\mathbf{R}_{\mathrm{BEV}} \in SO(2)$ and a translation vector $\mathbf{t}_\mathrm{BEV} \in \mathbb{R}^2$ are provided by RANSAC, allowing us to express the transformation between the BEV image pair $(\mathbf{I}_q(u,v), \mathbf{I}_r(u,v))$ as follows:
\begin{equation}
    \begin{aligned}
        \mathbf{R}_{\mathrm{BEV}}  &=
        \begin{pmatrix}
            \cos(\theta) & -\sin(\theta) \\
            \sin(\theta) & \cos(\theta)
        \end{pmatrix} \\\mathbf{t}_{\mathrm{BEV}}  &=
        \begin{pmatrix}
            t_u \\
            t_v
        \end{pmatrix}
        \end{aligned}
\end{equation}
\begin{equation}
\begin{aligned}
\mathbf{I}_q(u,v) &= \mathbf{I}_r(u',v')\\
\left[
        \begin{array}{c}
            u \\
            v
        \end{array}
        \right]
        = &\mathbf{R}_{\mathrm{BEV}}
        \left[
        \begin{array}{c}
            u' \\
            v'
        \end{array}
        \right]
        + \mathbf{t}_{\mathrm{BEV}}     
\end{aligned}
\end{equation}
Accordingly, the transform matrix $\mathbf{T}_{rq}$ between $\mathcal{P}^U_r$ and $\mathcal{P}^U_q$ is expressed as:
\vspace{-1mm}
\begin{equation}
    \mathbf{T}_{rq}=
    \begin{pmatrix}
    \cos(\theta) & \sin(\theta) & \mu tu \\
-\sin(\theta) & \cos(\theta) & \mu tv \\
0 & 0 & 1
    \end{pmatrix}
\end{equation}
where $\mu$ is the BEV image resolution since the translation is computed from BEV images. As the global pose $\mathbf{T}^G_r$ of the matched frame is stored in the database, we could obtain the global pose of $\mathcal{P}^U_q$ as:
\vspace{-1mm}
\begin{equation}
    \mathbf{T}_q=\mathbf{T}^G_r\mathbf{T}_{rq}
\end{equation}

\par \textbf{4) ICP Refinement.} 
The transform matrix $\mathbf{T}_{rq}$ obtained from BEV image matching serves as the initial guess for refinement using the ICP algorithm. This method performs fine alignment between the feature point clouds of the current keyframe $\mathcal{P}^U_q$ and the reference loop keyframe $\mathcal{P}^U_r$ like what \cite{BVMatch} does for place recognition. By optimizing the coarse transform, ICP ensures accurate registration and eliminates false positives, improving the robustness and precision of the loop closure process.
\par \textbf{5) Factor Graph Construction.} The refined transformation is then integrated into a factor graph as a loop closure factor. This, along with odometry and prior factors, facilitates optimize the graph to correct drift and maintain global consistency.
\vspace{-0.5mm}
\section{Experiment Results}
We evaluate our proposed BEV-LIO and BEV-LIO-LC methods on public datasets: Multi-Campus Dataset (MCD) \cite{MCD}, Multi-modal and Multi-scenario Dataset (M2DGR) \cite{M2DGR}, and Newer College dataset (NCD) \cite{NCD}. NCD and MCD's kth/tuhh sequences use handheld devices, MCD's ntu sequence employs an all-terrain vehicle (ATV), while M2DGR utilizes a ground robot. We compare our BEV-LIO with several state-of-the-art and widely used open-source methods, including FAST-LIO2, LIO-SAM, D-LIO, iG-LIO, and COIN-LIO. For loop closure evaluation, we compare our BEV-LIO-LC method with STD \cite{STD} integrated into FAST-LIO2, as well as FAST-LIO-SC and LIO-SAM-SC, both well-integrated by Kim et al. \cite{SC-LiDAR-SLAM} using \cite{ScanContext}. Due to the unavailability of certain closely related method BEV-LSLAM \cite{BEV-LSLAM} at the time of submission, we could not include them in our comparison. The experiments are conducted on an AMD Ryzen 7 6800H CPU and an NVIDIA GeForce RTX 3060 Laptop GPU running Ubuntu 20.04, using the REIN feature extractor from \cite{BEVPlace++} without additional training. The parameters of the experiments will be released later.



\vspace{-0.5mm}
\subsection{State Estimation Evaluation}
\vspace{-0.5mm}
\par \textbf{1) NCD \& MCD Results.} This section primarily evaluates methods using Ouster LiDAR dataset with different channels. The result of the Absolute Trajectory Error (ATE) of each method are reported in Table \ref{MCDAPE}. 
\par The cloister, math and park sequences in \cite{NCD} involve indoor-outdoor transitions and constrained spaces, while the under sequence is recorded underground. Due to these challenging conditions, iG-LIO failed in both the cloister and under sequences. In the cloister and math sequences, our proposed method achieves the best performance, while it performs second best in park and under sequences, demonstrating the potential of our BEV-LIO in geometrically challenging environments. Furthermore, our method achieves competitive results compared to COIN-LIO, which utilizes intensity images.

\begin{table*}[!t]
\setlength{\tabcolsep}{2.5pt} 
\caption{{\vspace{-0.5mm}} Absolute Trajectory Error of State Estimation Evaluation (RMSE, Meter)}
\label{MCDAPE}
  \vspace{-3mm}
  \begin{adjustbox}{max width=\textwidth}
  \begin{tabular}{lcccc ccccc ccccc}

\toprule
        Method        &Years  &cloister$^{\text{O*}}_{1}$       &math$^{\text{O*}}_{1}$        &park$^{\text{O*}}_{1}$        &under$^{\text{O*}}_{1}$       &ntu\_01$^{\text{O*}}_{2}$     &ntu\_02$^{\text{O*}}_{2}$      &ntu\_10$^{\text{O*}}_{2}$   &ntu\_04$^{\text{O*}}_{2}$    &ntu\_08$^{\text{O*}}_{2}$  &ntu\_13$^{\text{O*}}_{2}$   &kth\_06$^{\text{O}}_{2}$    &kth\_09$^{\text{O}}_{2}$  &kth\_10$^{\text{O}}_{2}$      \\ 
Dist.(m)& &428    &290   &2396   &236   & 3198   & 642   & 1783   & 1459   & 2421   & 1231   & 1403   & 1076   & 920                                                                                                                                                                     \\
        \midrule
        DLIO &2022   &0.150             &0.157             &0.458             &0.168             & 3.879            & 0.688            &4.659             &\underline{1.313} & 3.357            & 1.320            & 1.146            & 2.450            & 1.946                                \\
LIOSAM &2020&-                 &-                 &-                 &-                 & 2.603            & \textbf{0.277}   &2.570             &\textbf{0.736}    & 4.197            & 1.117            & 0.746            & 0.572            & 0.721                                      \\
        iG-LIO &2024&$\times$        &0.059           & \textbf{0.233} &$\times$        & 1.594          & 0.509          &1.811           &1.956           &\textbf{1.784}  & 1.257          & $\times$       & $\times$       &$\times$                                                    \\
        COIN-LIO&2024& \underline{0.062} & \underline{0.059} & 0.303             & \textbf{0.051}    & 6.450             & 0.772             & 3.054             & 2.370             & 7.076             & 1.909             &10.102             & 25.583            &13.571                 \\
        FAST-LIO2 &2021& 0.062             & 0.060             & 0.323             & 0.056             & \underline{1.414} & 0.416             &\underline{1.789}  &1.772              &1.959              & \underline{1.028} & \underline{0.615} & \underline{0.207} & \underline{0.718}       \\
        \textbf{BEV-LIO}&2025 & \textbf{0.053}   & \textbf{0.054}   & \underline{0.295}& \underline{0.052}& \textbf{1.396}   & \underline{0.410}& \textbf{1.618}   &1.627             &\underline{1.811} & \textbf{0.929}   & \textbf{0.580}   & \textbf{0.189}   & \textbf{0.367}           \\ 
        \toprule
        Method          &kth\_01$^{\text{O}}_{2}$     &kth\_04$^{\text{O}}_{2}$  &kth\_05$^{\text{O}}_{2}$ &tuhh\_02$^{\text{O}}_{2}$  &tuhh\_03$^{\text{O}}_{2}$ &tuhh\_04$^{\text{O}}_{2}$ &tuhh\_07$^{\text{O}}_{2}$ &tuhh\_08$^{\text{O}}_{2}$   &tuhh\_09$^{\text{O}}_{2}$  &gate\_03$^{\text{V}}_{3}$  &street\_01$^{\text{V}}_{3}$ &street\_03$^{\text{V}}_{3}$ &street\_04$^{\text{V}}_{3}$ &street\_08$^{\text{V}}_{3}$\\ 
        Dist.(m)          & 1416                          &1052   &919   &749   &1137   &297   &742   &1128   &290   &248    &752    &424    &840    &340    \\
        \midrule        
        DLIO              & 1.045                     & 1.277            & 3.103            & 0.921            & 2.284            & 0.685            & 1.380            & 1.571            &1.728             & 0.121            & 0.390            & \underline{0.141}& 0.629            & \textbf{0.136}   \\
        LIOSAM            & 12.745                & 0.506& 0.588            & 0.334            & 1.022            &\textbf{0.153}    & 37.205           & 37.735           & \textbf{0.125}   & \textbf{0.112}   & 0.527            & 0.144            & 0.928            & 0.204            \\
        iG-LIO            & $\times$                      & $\times$       & 3.791          & 6.825          & 3.635          & 4.340          &$\times$        & 5.995          & 0.582          & 0.193          & 0.383          & \textbf{0.125} & 0.517          &0.185           \\ 
        COIN-LIO2&17.134  &24.597                      &16.078             &14.493             &39.386             &3.011              &3.584              &6.758              &3.867              & -                 & -                 & -                 & -                 & -                 \\
        FAST-LIO2 & \underline{0.568} & \underline{0.476}           & \underline{0.574} & \underline{0.209} & \underline{0.800} & \underline{0.167} & \textbf{0.244}    & \underline{0.716} & \underline{0.156} & 0.214             & \underline{0.286} & 0.184             & \underline{0.434} & 0.231        \\      
        \textbf{BEV-LIO}  & \textbf{0.363}                & \textbf{0.289}   & \textbf{0.488}   & \textbf{0.203}   & \textbf{0.676}   & 0.168            & \underline{0.253}& \textbf{0.643}   & 0.156            & \underline{0.121}& \textbf{0.285}   & 0.149            & \textbf{0.418}   &  \underline{0.181}  \\

  \bottomrule
  \end{tabular}
  \end{adjustbox}
    \begin{tablenotes}
    \item[1] 1.$\times$ denotes a complete failure of the method. - indicates that LIO-SAM requires 6-axis IMU data, or COIN-LIO cannot run without Ouster LiDAR. 
    \item[2] 2.\textbf{Bold} indicates the best result, and \underline{underline} indicates the second-best result.
    \item[3] 3.The notations $^{\text{O*}}$, $^{\text{O}}$ and $^{\text{V}}$ and represent the OS1-128 LiDAR, OS1-64 LiDAR, and VLP-32C LiDAR respectively. The subscripts $_{1}$, $_2$ and $_3$ correspond to the datasets NCD, MCD, and M2DGR.
\end{tablenotes}
  \vspace{-6mm}
  \end{table*}

\par
\begin{figure*}[!ht]
\centering
\includegraphics[width=\linewidth]{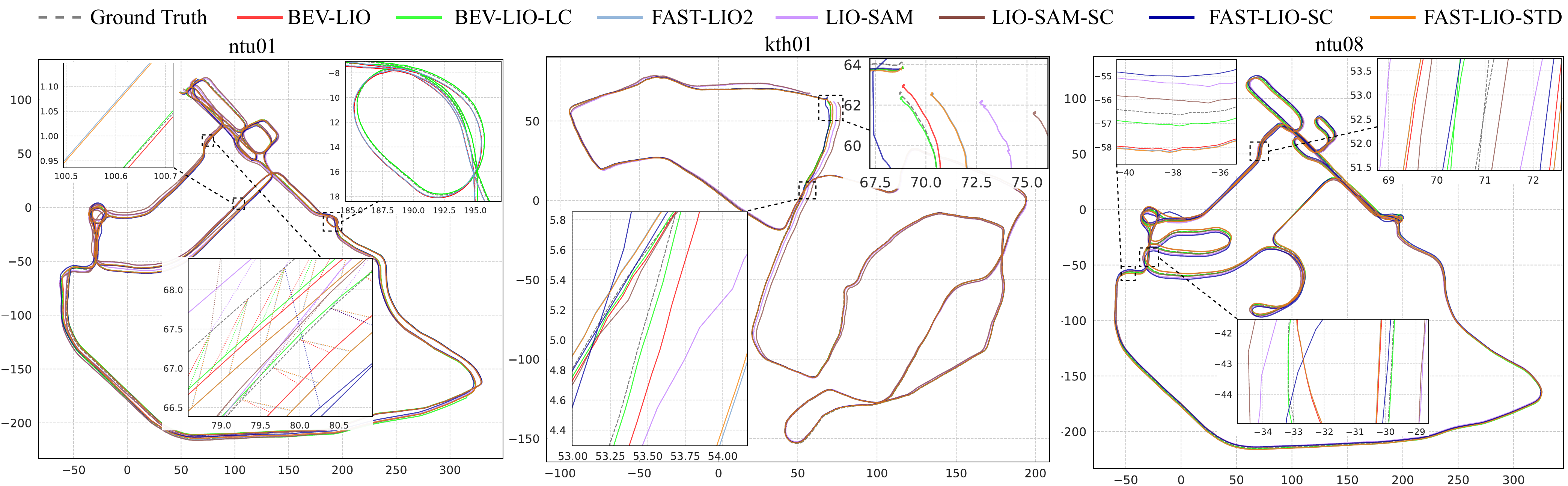}
\caption{\vspace{-4mm}Trajectory comparison between LIO and LC methods and the zoom-in views in several sequences.}
\label{exam_loop}
\vspace{-7mm}
\end{figure*}

 In the ntu sequences, all methods exhibit stable performance due to the ATV platform's reliable measurements. Except ntu\_04, BEV-LIO performed the best or the second best. Despite utilizing the high-cost OS1-128 LiDAR, COIN-LIO falls short as it is only impressive when dealing with long corridor with texture. As illustrated in Fig. \ref{exam_loop}, the trajectory error of BEV-LIO in the upper left zoom and middile zoom of ntu\_01 is noticeably smaller than that of FAST-LIO2.

\begin{figure}[htbp]
\vspace{-1mm}
\includegraphics[width=\linewidth]{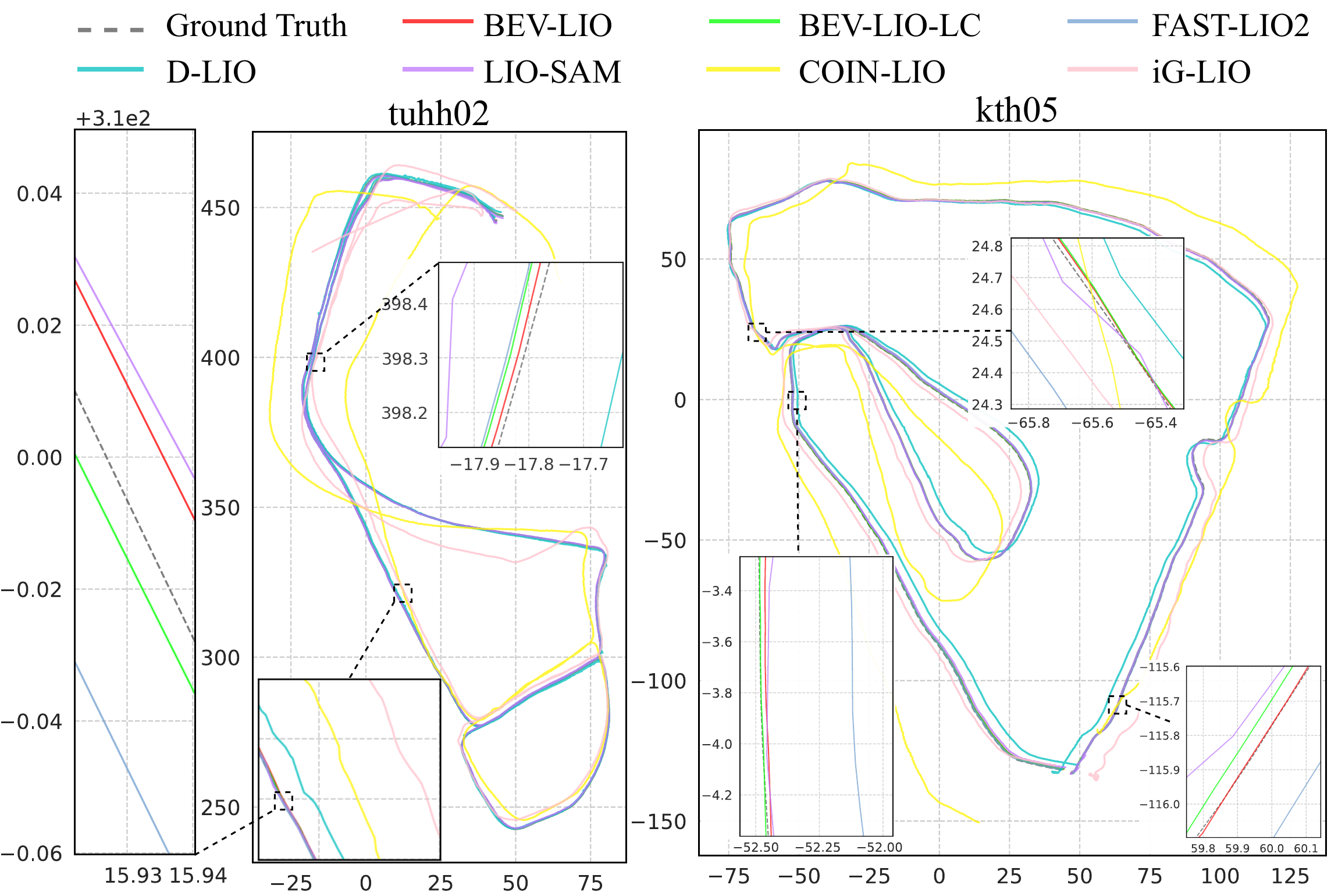}
\caption{\vspace{-0.1mm}Trajectory comparison in kth\_05 and tuhh\_02 sequences.}
\label{exam_xiao}
\vspace{-8mm}
\end{figure}

\par For the kth and tuhh sequences, BEV-LIO achieved the best results and performed better than FAST-LIO2 besides 3 sequences in tuhh, which proves the effectiveness of our BEV part. With a handheld setup, the recorded sequences may experience more pronounced shaking than those captured with an ATV, potentially leading to some methods failure or significant drift in the kth and tuhh sequences. Due to the use of the OS1-64 LiDAR in the kth and tuhh sequences, COIN-LIO did not achieve the expected performance. 
Our method achieves an average ATE improvement of 29.3\% over FAST-LIO2 for the kth sequences. Notably, BEV-LIO reduces ATE by 48.9\% in kth\_10, 49.1\% in kth\_04, highlighting the effectiveness of BEV features in challenging environments. As shown in Fig. \ref{exam_loop}, BEV-LIO demonstrates a good alignment with the ground truth in the kth\_01 sequence without a back-end optimization. Notably, in the upper right zoom, which corresponds to the endpoint of the sequence, our method’s trajectory is over 0.1m closer to the ground truth than FAST-LIO2. The kth\_05 and tuhh\_02 sequences in Fig. \ref{exam_xiao} clearly demonstrate that our BEV-LIO outperforms the compared methods, exhibiting better alignment with the ground truth with our BEV components.

\par \textbf{2) M2DGR Results.} This part primarily focuses on Velodyne LiDAR, demonstrating the capability of BEV-LIO across different types of LiDAR setups. We select several sequences from M2DGR. As shown in Table \ref{MCDAPE}, our method has lower ATE compared to FAST-LIO2 in all sequences. Even if BEV-LIO isn't the best result, our method errors differ from the best methods by only 0.01 m to 0.05 m and reduces FAST-LIO2's error by 0.09m in gate\_03 sequence and 0.05m in street\_08 sequence.
\par \textbf{3) Runtime Analysis.} To evaluate the runtime performance of our method, we tested it in the clositer sequence, where our approach achieves an average processing time of 62.7 ms per frame (16 Hz) on our laptop, with the hardware configuration as mentioned earlier.



\begin{table*}[!ht]
\setlength{\tabcolsep}{3pt} 
  \caption{\vspace{-0.8mm}Loop Method Absolute Trajectory Error (RMSE) (\textit{m}) / Improvement (RMSE) (\textit{m}) }
  \vspace{-3mm}
  \begin{adjustbox}{max width=\textwidth}
\begin{tabular}{lcccc ccccc ccc}

  \toprule
  Method &  ntu01    & ntu08 & kth06 & kth09 & kth01 & kth04 & kth05 & tuhh02  & tuhh08  \\ 
  \midrule
\rowcolor{gray!20} 
\textit{LIO Methods}&&&&&&&&&\\

  LIO-SAM   
  &2.603   
  & 4.197 & 0.746 & 0.572 
  & 12.745& 0.506 & 0.588 
  &{0.334}    & 37.735  \\

FAST-LIO2  &1.414    & 1.959 & 0.615 & 0.208 & 0.476 & 0.568 & 0.574 & 0.209    & 0.716   \\

  \textbf{BEV-LIO}       &1.396    & 1.811 & 0.580 & 0.189 & 0.363 & 0.289 & 0.488 & 0.203    & 0.643   \\
  
\rowcolor{gray!20} 
\textit{LIO-LC Methods}&&&&&&&&&\\

  LIO-SAM$^*$ 
  &3.840/$+$1.237  
  & 3.700/\textbf{$-$0.497} & 0.725/$-$0.021 & 1.570/$+$0.998
  & 10.539/\textbf{$-$2.206}& 0.302/{$-$0.204} & \textbf{0.355}/\textbf{$-$0.233}
  & 0.415/$+$0.081  & 26.788/\textbf{$-$10.947}  \\
  
LIO-SAM-SC   &3.516/$+$0.913 &3.757/$-$0.440 &0.784/$+$0.038  &1.136/$+$0.564  &13.871/$+$1.126  &0.268/$-$0.238  &2.528/$+$1.954   &0.427/$+$0.093  &36.917/$-$0.818 \\

FAST-LIO-SC    &2.537/$+$1.123  &2.593/$+$ 0.634    &1.301/$+$0.686    &2.332/$+$2.125   &0.886/$+$0.410    &1.542/$+$0.974     &0.525/$-$0.049  &0.947/$+$0.738   &0.726/$+$0.010     \\

FAST-LIO-STD   &1.417/$+$0.003  &1.964/$+$0.005   &0.617/$+$0.002                     &0.208/$+$0.001                      &0.473/$-$0.003  &\textbf{0.199}/\textbf{$-$0.369}   &0.552/$-$0.022  &0.211/$+$0.002   &0.720/$+$0.004     \\
  \textbf{BEV-LIO-LC}  &\textbf{1.323}/\textbf{$-$0.073}    & \textbf{1.402}/$-${0.409} & \textbf{0.492}/\textbf{$-$0.088} & \textbf{0.187}/\textbf{$-$0.002}
   & \textbf{0.347}/$-$0.016 & {0.214}/$-$0.075 & {0.439}/$-$0.049 & \textbf{0.199}/\textbf{$-$0.004}     & \textbf{0.640}/{$-$0.003}   \\
  
  \bottomrule
  \end{tabular}
  \end{adjustbox}
  \label{LOOPAPE}
  \vspace{-2mm}
  \end{table*}

\vspace{-0.5mm}
\subsection{Loop Closure Evaluation}
\vspace{-0.5mm}
In this section, we evaluate the performance of our BEV-LIO-LC in comparison with LIO-LC methods. The ATE results are reported in Table \ref{LOOPAPE}, accompanied by the improvement achieved through the loop closure algorithm. 
\par As shown in the left zoomed-in views of ntu\_01 in {Fig. \ref{exam_loop}}, our BEV-LIO-LC aligns closer to the ground truth (0.1m to 0.5m) than other LC methods. Particularly noteworthy is the right zoomed-in view, where the path forms an elliptical shape and multiple passes. Our method accurately fits the ground truth, reducing the error by nearly 0.5m compared to FAST-LIO2 and our front-end BEV-LIO. Similarly, in kth\_01, our method better converges to the ground truth at the path’s endpoint (as shown in the right zoom-in view), demonstrating the global consistency and accuracy of our loop closure algorithm. As shown in Table \ref{LOOPAPE}, our approach exhibits superior ATE performance across 77.8\% selected sequences compared to the results in Table \ref{MCDAPE}. Notably, for the ntu\_08 sequence, our method achieves an ATE reduction of 0.409m, representing a significant improvement of 22.6\% as illustrated in Fig. \ref{exam_loop}. The local zoom-in views of the ntu\_08 sequence reveal that our BEV-LIO-LC effectively corrects the trajectory at loop closure locations, even when the front-end estimates are less accurate. Our loop closure module is capable of refining the trajectory to outperform other LC methods, resulting in corrections ranging from over 0.5m to more than 1m. In Table \ref{LOOPAPE}, our method consistently reduces the ATE, whereas other LC methods result in an increase in ATE. Although LIO-SAM$^*$ achieves a greater reduction in ATE in several sequences compared to our BEV-LIO-LC, it ultimately exhibits a significant drift, failing to correct the cumulative errors from the front-end. In contrast, BEV-LIO-LC demonstrates superior stability and effectiveness.

\par In summary, our method demonstrates superior performance compared to direct point cloud matching methods like \cite{DLIO} and spherical projection methods like \cite{COIN-LIO}. While they perform well in certain conditions, BEV-LIO(LC) achieves higher localization precision by coupling BEV image feature and a back-end optimization. This highlights the advantage of our method in improving localization accuracy and loop closure performance for correcting accumulated errors. 

    \vspace{-0.5mm}
\section{Conclusion}
    \vspace{-0.5mm}
    
In this work, we propose BEV-LIO(LC), a LIO framework that combines BEV images of LiDAR data with geometric registration for front-end odometry and integrates loop closure via BEV image features and factor graph optimization for the back-end. Our approach enhances front-end odometry by combining reprojection error from BEV image matching with geometric registration, and utilizes iEKF to couple these residuals, improving odometry accuracy and reliability. For loop closure, our back-end system employs a KD-tree-indexed BEV descriptor database. Upon loop detection, a coarse transform computed via RANSAC from BEV image matching initializes the ICP process, which refines the transform for improved global consistency of localization when integrated into a factor graph alongside odometry factors. Extensive experiments across various scenarios and with different LiDAR validate the competitive localization accuracy of BEV-LIO(LC) compared to state-of-the-art methods. Looking ahead, future work will focus on eliminating the dependency on CNNs for feature extraction, with the goal of further enhancing real-time performance of the system.

    \vspace{-2mm}
\bibliographystyle{IEEEtran}
\bibliography{citation/citations} 

\end{document}